\documentclass{article} 

\usepackage[utf8]{inputenc} 
\usepackage[T1]{fontenc} 

\usepackage{pdfpages}
\usepackage{svg}
\usepackage{latexsym, amsmath, amssymb, amsfonts, amsthm}

\usepackage[size=tiny, backgroundcolor=white]{todonotes}

\usepackage{enumitem}
\usepackage{url}
\usepackage{hyperref}
\usepackage{physics}
\usepackage{mathtools}
\usepackage{multicol}
\usepackage{tikz-cd}
\usepackage{float}

\usepackage[left=2cm,right=2cm, top=2cm,bottom=2cm]{geometry}

\newtheorem{thm}{Theorem}[section]
\newtheorem{lem}[thm]{Lemma}
\newtheorem{cor}[thm]{Corollary}

\newtheorem{example}[thm]{Example}
\newtheorem{remark}[thm]{Remark}

\newtheorem{definition}[thm]{Definition}
\newtheorem{open}[thm]{Open problem}
\theoremstyle{plain}

\title{Comparison of different Unique hard attention transformer models by the formal languages they can recognize} %
\author{Leonid Ryvkin}

\begin{document}

\maketitle

\begin{abstract}	
This note is a survey of various results on the capabilities of unique hard attention transformers encoders (UHATs) to recognize formal languages. We distinguish between masked vs. non-masked, finite vs. infinite image and general vs. bilinear attention score functions. We recall some relations between these models, as well as a lower bound in terms of first-order logic and an upper bound in terms of circuit complexity.
\end{abstract}

\begin{center}

\includegraphics[width=0.4\textwidth]{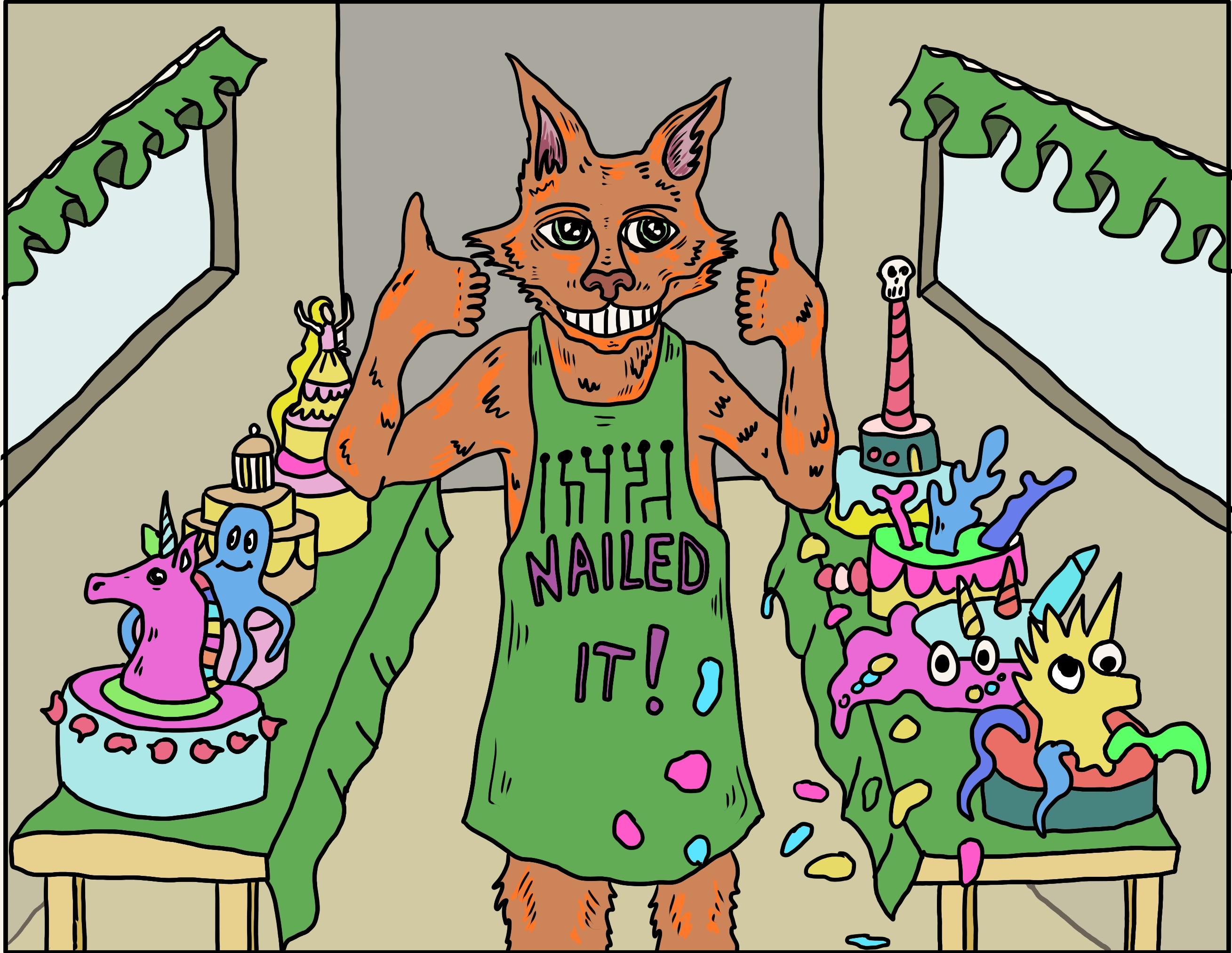}  \\  
{\tiny Image credit: Anna Marklová}

\end{center}

\tableofcontents 

\section{Introduction}
The goal of this note is to give an overview of the capabilities of different flavors of unique hard attention transformer encoders in terms of the formal languages they are able to recognize. This study is relevant in the context of the rising use of large language models, which typically follow a transformer architecture. While the model we will be primarily investigating has features very distinct from real-world transformers (we will comment on the distinction later) they can still give valuable insights into the principle underlying transformer capabilities.\\

Roughly speaking, a transformer can be thought of function that, given an input of any length, can construct a sequence of the same length. It transforms one sequence into the other. Typically, a transformer can be seen as a concatenation of two types of operations (usually referred to as layers):
\begin{itemize}
	\item Point-wise operations, i.e. operations where the $i$-th position of an input sequence is transformed to the $i$-th position of the output sequence.
	\item Attention operations, i.e. those where the function at position $i$ has some access to input sequence elements other than $i$.  
\end{itemize}
To obtain a reasonable computational model the type of access to other positions that we are allowing will be significantly restricted. The first type of operation, however, for all practical purposes can be assumed completely arbitrary: It is usually realized by feed-forward neural networks which can approximate any (continuous) function. Moreover, in the scenarios we consider, we will usually be able to forget about those operations all-together, because they can be simulated by attention operations (where we just don't care what other sequence positions are involved and don't use the data they provide).\\

The  types of transformers we will focus on are those with unique hard attention: At a position $i$, their attention operations compute a score between the $i$-th sequence element and the $j$-th one (for all $j$) and the final calculation result may only depend on the value at the $i$-th position and at the unique position having scored highest. If the highest score is not unique, a tie-breaking strategy like taking the earliest or latest element with the highest score will be employed. Depending on which masking is used, this will have an effect on the transformers capabilities.\\

Another tie-breaking strategy would be taking the average hard attention, where ties are resolved by taking the arithmetic average of the highest-scoring values. This apparently small change has a huge effect on expressiveness of transformers \cite{merrilsat2021}: Despite being closer to capabilities of actual transformers (which utilize an even more different attention mechanism called softmax), they are very hard to study from a viewpoint of formal logic. Unlike for unique hard attention, the average hard attention transformers impose very few limitations on the information flow of the function they induce: If all scores are identical, then the $i$-th position of the next layer can depend on all positions of the previous one.\\

Our way of analyzing transformers will be by comparing which classes of formal languages they can recognize: A transformer class being able to recognize more languages will be considered more expressive. In order to allow a transformer to recognize a language additional data is needed, e.g. encoding some information on the positions of the letters into the input sequence (or describing how the output sequence indicates belonging to the language or not.) The choice of allowed encoding will have a significant impact on the transformers expressiveness.\\

Sometimes we restrict the range of $j$'s considered, e.g. to the ones which came before $i$ in the sequence chronologically. This is called masking and its effect will be discussed in detail. Somewhat surprisingly, on less expressive models (where the initial information has only a bounded number of bits), masking is actually increasing expressiveness: It permits to have some additional information on the order of sequence elements. For more expressive models (where the number of  input bits per sequence element might be increase with the length of the sequence) this advantage disappears: In the most powerful  setting, non-masked transformers suddenly can do everything (and maybe more) than masked ones.\\

\subsection{Literature overview}
Our discussion is mainly based on the following articles:
\begin{itemize}
	\item \cite{hahn-2020-theoretical}. This paper established a first important limitation of the expressiveness of generalized unique hard attention transformers (GUHATs): The fact that all languages described by them have to be (strongly )$\epsilon$-fixable, i.e. that by dictating a small part of their characters one can either force a string to be in the language or not. 
	\item \cite{haoFormalLanguageRecognition2022}. This paper developed a further upper bound for GUHATs in terms of circuit complexity: They showed that any GUHAT can be simulated by an $AC^0$ circuit family, thus languages not recognizable by these circuits can not be recognized by transformers.  
	\item \cite{barceloLogicalLanguagesAccepted2023}. This article showed that unique hard attention transformers can recognize all languages definable by first order logic with unary numerical predicates. For this it used linear temporal logic, which has later been utilized in \cite{yangMaskedHardAttentionTransformers2024}. Moreover, it contained the observation that GUHATs can not recognize the language $\mathtt{APPROXIMATE- MAJORITY}$ and hence can not express all languages from $AC^0$, showing that the bound from \cite{haoFormalLanguageRecognition2022} is not sharp.
	\item \cite{yangMaskedHardAttentionTransformers2024} investigated GUHATs, for which the image of the positional encoding may only have finitely many bits. They showed that on this level, the masking is actually an advantage and showed that the masked versions recognize exactly the languages definable by first order logic with unary numerical predicates. They also provided precise characterizations of certain, even less expressive, classes of transformers.
\end{itemize}
Some of the results we quote were presented in the lecture \cite{kozachinskiy2023logical} available on YouTube, but have not yet been published. Furthermore, we profited from the survey \cite{weissThinkingTransformers2021}.

\subsection{Potentially new observations}

\begin{itemize}
	\item The proof of Theorem \ref{thm:guhatdepthisimportant}, stating that the expressiveness of GUHATs can be increased by adding layers is, to our knowledge, new. It has similarities with depth hierarchy results of \cite{yangMaskedHardAttentionTransformers2024}, and follows from considerations of \cite{yao-etal-2021-self} and \cite{haoFormalLanguageRecognition2022}.
	
	\item The notion of separable attention  (cf. Definition \ref{def:sep}), while implicitly attacked in multiple proofs and examples in the literature, was not yet explicited as a way to describe the difference between GUHAT and UHAT transformers. We use this notion to show that attention functions of GUHATs whose embeddings have finite image are always separable (cf. Theorem \ref{lem:finitelyalliseparable}), thus conceptually explaining why the difference between GUHAT and UHAT disappears in the context of finite image embeddings.

	\item The formal proof of Theorem \ref{thm:fmaskedbetter}, showing that in the context of finite image UHATs the masked version can simulate the non-masked one seems to be new. While \cite{yangMaskedHardAttentionTransformers2024} discusses the effect on masking, this result does not seem proven/described there. 
\end{itemize}

\subsection{Acknowledgements}
This note is an abridged version of the author's master thesis 
\emph{Capabilities and limitations of Transformer Encoders} under the supervision of {Prof. Thomas Zeume} and {Dr. Nils Vortmeier} at the University of Bochum. I would like to thank Thomas Zeume for proposing this very interesting topic and insightful discussions. I also would like to thank Corentin Barloy and Nils Vortmeier for their support, explanations and generous contribution of new ideas and questions. I would furthermore like to extend my thanks to Pablo Barcelo and Alexander Kozachinskiy for a very helpful email exchange.

\section{Variants of unique hard attention transformer encoders}
In this section, we will introduce multiple classes of unique hard attention transformers and show that the most expressive of this classes adheres to  a depth hierarchy in terms of language recognition (i.e. that expressiveness can be increased by adding layers). \\

Instead of working with unique hard attention transformer encoders (UHATs) directly, we will use the equivalent but more abstract  URASP (unique hard attention restricted access sequence programming) language.

\subsection{GUHAT transformers without masking}

The RASP (Restricted access sequence programming) language was introduced in \cite{weissThinkingTransformers2021} in order to give a programmatic style to define transformers. We present here an adapted and simplified version, similar to \cite{yangMaskedHardAttentionTransformers2024}, which describes exactly GUHAT type transformers (as described in \cite{haoFormalLanguageRecognition2022}).

\begin{definition}A URASP (Unique hard attention RASP) program is given by a finite sequence of operations on lists. The initial input (line $l=0$) is given by an initial list $L_0$, the entries of which are in an (a priori arbitrary) set $V[0]$). The entries subsequent lines have will be drawn from further sets $V[i]$ at line $i$.  
	
	\begin{itemize}
		\item A point-wise operation (in line $l$) is any function $f$ having at position $i$ access to the values $L_0(i)$,...,$L_{l-1}(i)$. If $V[0],...,V[l-1]$ are the types of the first $l$ lines, then this is exactly a function $R:V[0]\times...\times V[l-1]\to V[l]$. We write $L_{l}(i)=R(i)$, actually meaning $R(i)=R(L_0(i),...,L_{l-1}(i))$.
		\item An attention operation (in line $l$) is written as follows $$ L_{l}(i)=\blacktriangleright_j[s(i,j)] v(i,j)$$
		It contains the following information:
		\begin{itemize}
			\item $ \blacktriangleright_j$ means that we use rightmost hard attention, i.e. that in case of scoring ties the rightmost value for $j$ is chosen.
			\item $[s(i,j)]$ is an attention function. It is given  by $$s_l:V[0]\times...\times V[l-1]\times V[0]\times...\times V[l-1]\to \mathbb R.$$
			It yields a matrix: for every $(i,j)$ it computes the attention score of column $i$ with column $j$. The letter $s$ stands for score.
			\item $v(i,j)$ is an activation function
			$$v_l:V[0]\times...\times V[l-1]\times V[0]\times...\times V[l-1]\to  V[l].$$
			The letter $v$ stands for value.
		\end{itemize} 
		The output of such a line is by computing the attention matrix, then picking the maximal $j$ for each $i$ and evaluating $v$ on these couples. Ties are resolved favouring the right ( $ \blacktriangleright$).
	\end{itemize}
\end{definition}

\begin{remark}
 The point-wise operations are, strictly speaking, unnecessary: They can be simulated by attention operations. (one picks an arbitrary attention function and makes the activation function independent of the $j$ values).
\end{remark}
\begin{definition} We say that a language $L$ over an alphabet $\Sigma$ is recognizable by a generalized unique hard attention transformer encoder  (GUHAT) if there is an initialization $\{E_n:\Sigma\times \{0,...,n-1\}\to V[0]~|~n\in\mathbb N\}$, URASP program (with first line $V[0]$ and $d$ lines) and acceptance set $valid\subset V[d]$. such that a string $w\in\Sigma^*$ is in $L$ if and only if applying the URASP program to $w$ leads to a list $L_d$ such that $L_d(n-1)\in valid$. 
	
We call \emph{GUHAT} the class of languages which can be recognized by GUHATs.
\end{definition}

\begin{example}[adapted from \cite{kozachinskiy2023logical}] \label{ex:palin}
	Let us define a URASP program recognizing $\mathtt{PALINDROMES}$ over $\Sigma$\footnote{i.e. the language of strings that read the same forwards and backwards over an alphabet which has at least two characters.}. We do it layer by layer:
	\begin{enumerate}[start=0]
		\item Given an input ($w_0,...,w_{n-1}$), we initialize the list by the triples $$(w_0,0,n),...(w_{n-1},n-1,n).$$
		\item We perform an attention operation $v$  with $s(i,j)$ defined as follows:
		$$
		s_1((c,i,n),(\tilde c,j, n))=-(n-1-i-j)^2
		$$
		This value is clearly maximized for $i=n-1-j$. We then perform the activation:
		$$
		v_1((c,i,n),(\tilde c,j, n))=\begin{cases}
			1& \mathrm{if} ~c=\tilde c\\
			0 & \mathrm{else}
		\end{cases}
		$$		
		Hence after this step in addition to having $l_0(i)=(c,i,n)$ we have a fourth value $l_1(i)\in\{0,1\}$.
		\item In this step we apply the attention operation with:
		$$s_2((l_0(i),(l_1(i)),(l_0(j),(l_1(j)))=-l_1(j).$$
		This will find a $j$ with $l_1(j)=0$ if such a value exists. We then set
		$$v_2((l_0(i),(l_1(i)),(l_0(j),(l_1(j)))=l_1(j),$$
		i.e. this zero if and only if for \emph{some} $j$ we have $l_1(j)=0$, else it is 1.
	\end{enumerate}
	We pick the subset $valid=\{1\}\subset V[2]=\{0,1\}$. By construction this recognizes $\mathtt{PALINDROMES}$.
\end{example}

\begin{remark}\label{rem:theirguhatourguhat}Our formulation differs slightly from the one of \cite{haoFormalLanguageRecognition2022}. We are formulating everything in terms of the URASP language, rather than an transformer. Moreover, there are a few additional technical choices, which do not change the expressivity. Let us highlight them:
	\begin{itemize}
		\item We are talking about the right-most hard ($\blacktriangleright$) tie resolution strategy. We could have formulated everything with the leftmost choice in case of ties (we will use $\blacktriangleleft$ to indicate that). This distinction does not change anything: Any URASP program with initialization recognized the same languages as one, where no ties occur. This is shown in the proof of \cite[Proposition 1]{haoFormalLanguageRecognition2022}. To do this, one records for every $n$ the smallest distance $\epsilon_n$ which ever occurs between two attention values when the URASP program operates on strings of length $n$. Then one replaces $E_n$ with an initialization, which contains the current position $i$ and the input length $n$, so one can use these values in all attention and activation functions. Then one changes all the attention functions $s(i,j)$ to $s(i,j)+j\cdot \frac{\epsilon_n}{2}$. This way in case of identical values the rightmost term will be favoured, and due to the choice of $\epsilon_n$ orders of distinct scores are never changed.
		\item Our programs compute only one score and have attention only towards one value on each level, while in \cite{haoFormalLanguageRecognition2022} multiple attention heads are allowed. This does not change expressivity, since we can simulate $H$ attention heads by having $H$ consecutive lines with one attention function each.
	\end{itemize}
\end{remark}

We will be interested in two additional properties:

\begin{definition}\label{def:uhat} A language is in \emph{UHAT} (resp. can be recognized by a UHAT), if there exists a URASP program with the properties
	\begin{itemize}
		\item The $V[l]$ are finite-dimensional vector spaces $\mathbb R^{k_l}$,
		\item The attention maps $s_l$ are bilinear,
	\end{itemize} 
which recognizes it.
\end{definition} 

\begin{remark}\label{rem:vspacesimple} To make the setting more realistic, one could require that the activation functions and the pointwise operations can be computed by feed-forward neural networks.
However for every fixed $n$ the number of inputs to the program is finite, so also the number of possible score/ activation values is. Taking the union over all $n$ we still have a countable set, which can be embedded into an $\mathbb R^{k_l}$ in a discrete manner. But any function on a discrete set can be realized by a feedforward neural network.\\
For the same reason the restriction of the spaces to be  $\mathbb R^{k_l}$ is essentially meaningless and just there so bilinearity makes sense. We will get rid of this technicality in Section \ref{sec:sep}
\end{remark}

\begin{definition}\label{def:fuhat} We say that a URASP program with initialization $\{E_n,n\in\mathbb N\}$ has finite type, if $\bigcup_n \mathrm{Image}(E_n)$ is finite. We call the class of languages recognizable by such a program \emph{F-GUHAT} (respectively \emph{F-UHAT}, if it also has bilinear attention functions).
\end{definition} 

\begin{remark}
For a fixed $n$, the finiteness of the image of $E_n$ is automatic, but for the union of all of them this is a real restriction: For instance the length of the current word $n$ may not be encoded by $E_n$, meaning that e.g. the trick to show that attention score ties can be avoided from Remark \ref{rem:theirguhatourguhat} can not be carried out any more.
\end{remark}

\subsection{GUHAT transformers with masking}

The idea of masking is that the attention mechanism of a transformer does not have access to all values, but only to a certain subset of them, the remaining values are \emph{masked}. In this context a masking is a family of functions $$\{m=m_n:\{0,...,n-1\}\times \{0,...,n-1\}\to \{0,1\}~|~n\in \mathbb N\}$$
such that the attention operation at position $i$ of the URASP program only chooses among the $j$ with $m(i,j)=1$. We will be only interested in the following masking functions:
\begin{itemize}
	\item no masking $m(i,j)=1$ for all $i,j$.
	\item strict future masking $m(i,j)=1$ if and only if $i>j$.
		\item strict past masking $m(i,j)=1$ if and only if $i<j$.
\end{itemize}

With masking, it can happen that the set of attention scores we are maximizing over is empty (e.g. in the case of strict future masking, at the first position $i=0$). In this case we need to provide a default value which the calculation can take. (This value may depend on the values in the current column $i$.) Hence, following the approach used for BRASP in \cite{yangMaskedHardAttentionTransformers2024}, we will slightly extend the specification of attention operations for URASP: 
$$
L_{l}(i)=\blacktriangleright_j[m(i,j),s(i,j)]v(i,j):D(i)
$$
where  $m(i,j)$ is the masking and $D$ is the default value calculated on the current column. It behaves like a pointwise operation, i.e. at level $l$ it is a function:
$D:V[0]\times...\times V[l-1]\to V[l].$

\begin{remark} When we use masking, our tie resolution strategy starts being important. We will usually stick to strict-future masking and combine it with rightmost-hard attention (i.e. ties always being resolved to the right), however in the proof of Theorem \ref{thm:fmaskedbetter} we will temporarily need strictly past-masked layers combined with leftmost hard attention, which will then be denoted by $\blacktriangleleft$ in the syntax. (cf. also Remark \ref{rem:pastfutre}).
\end{remark}

\begin{definition} We say that a language is in \emph{MGUHAT} if it can be recognized by a URASP program with rightmost-hard attention and strict future-masking. Similarly we define \emph{MUHAT}, \emph{F-MUHAT}, \emph{F-MGUHAT} when the restrictions from Definitions \ref{def:uhat} and \ref{def:fuhat} are also applied.
\end{definition}

\begin{remark}In the case of strict future-masking information can only propagate from left to right. This means that the choice of position on which we check the validity of a string, the last position in our case, becomes significant: If we would read validity at $i=0$, we would significantly reduce the number of recognizable languages. Conversely, if we were to use past-masking, we would need to evaluate at position $i=0$ rather than $n-1$ in order to get maximum expressivity. 
\end{remark}

\begin{remark}\label{rk:maskedguhat}
For the GUHAT case, simulating masking by a non-masked transformer is very easy: Without loss of generality, the current position $i$ and input length $n$ are present in the initialization. Let $s_k$ be a given attention function and $m_n$ the masking we want to simulate. For any fixed $n$ the possible input sequences are a finite set, hence the set of values that $s_k$ can take are bounded from below by some number $K_n$. We can now set $$s_k'((v,i,n),(w,j,n))=     \begin{cases}
	s_k((v,i,n),(w,j,n)) & m_n(i,j)=1\\
	-K_n-1 & \rm{else}
\end{cases}$$
Hence, the values forbidden by the masking will never be the highest possible values. For positions, where with masking we would be aggregating over the empty set, we would need to modify the corresponding activation function by hand to compute the corresponding $D(i)$. In particular we have $MGUHAT\subset GUHAT$. 
\end{remark}

Let us adapt Example \ref{ex:palin} to the masked setting:

\begin{example}\label{ex:mpalin}
Let us adapt Example \ref{ex:palin} to the strictly future-masked setting. To do so, we observe that the original UHAT we constructed was doing all of its work twice: It was both comparing whether the letter at position $i$ coincides with the letter at position $n-1-i$  and vice-versa. Hence to construct the strictly future-masked UHAT, we can pick exactly the same attention function, but we change the first layer activation from:
	$$
	v_1((c,i,n),(\tilde c,j, n))=\begin{cases}
		1& \mathrm{if} ~c=\tilde c\\
		0 & \mathrm{else}
	\end{cases}
	$$
	to 
	$$
	\tilde v_1((c,i,n),(\tilde c,j, n))=\begin{cases}
		1& \mathrm{if} ~c=\tilde c \mathrm{~or~} i< \frac{n-1}{2}\\
		0 & \mathrm{else}
	\end{cases}.
	$$
	This way all positions before the middle, do not provide any information (they get the value 1 in any case), and we are checking whether a character $w_i$ is equal to $w_{n-1-i}$ at the position $\max\{i,n-1-i\}$ only.
\end{example}

\section{A depth hierarchy}
In this section, we will show that the expressivity of GUHATs can always be increased by adding layers (i.e. increasing the number of allowed lines in the corresponding URASP program). A similar theorem has been proved in \cite{yangMaskedHardAttentionTransformers2024} in the context of finite-image transformers.

\begin{thm}\label{thm:guhatdepthisimportant}
	For any $D\in \mathbb N$, there exists a $D'>D$ and a language that can be recognized by a GUHAT with $D'$ layers, but not by a GUHAT with $D$ layers.
\end{thm}

\begin{proof}
	We proceed by contradiction: Assume there would be a $D$ such that all languages recognizable by a GUHAT can be recognized by a GUHAT with $D$ layers. This would in particular mean that the languages $\mathtt{DYCK}-(1,D')$ (which are recognizable by a GUHAT by Example \ref{ex:dyck}) can for all $D'$ be recognized by transformers (i.e. URASP programs) with spaces/attention functions and activation functions $(V^{D'},s^{D'},v^{D'})$ and only $D$ layers (i.e. program lines). We claim that no such family of transformers can exist, because from such a family we would be able to construct a $D$-layer transformer $(\bar V,\bar a,\bar D)$ recognizing $\mathtt{DYCK}-1$ (which is not recognizable by a GUHAT by Example \ref{cor:dycknotac}). 
	We will make heavy use of the fact, that a URASP program is allowed to have completely different behavior for each input length $n$. Let us for simplicity assume that each $V[i]$ contains the value $0$.
	\begin{itemize}
		\item $\bar V[i]=\mathbb N\times \Pi_{D'\in\mathbb N} V^{D'}[i]$, i.e. an element of $\bar V[i]$ is just an infinite tuple of elements, where the first is a natural number and the others are just the spaces from the individual transformers from the given family. 
		\item The initialization $\bar E$ is given as:
		$$
		\bar E_n(\sigma,i)=\begin{cases}
			n& \mathrm{in~the~}\mathbb N-\mathrm{component}\\
			0 & \mathrm{in~the~}V^{D'}[0]\mathrm{-components~with~} D'\neq n\\
			E^n_n(\sigma,i)& \mathrm{in~the~} V^{n}[0]\mathrm{-component}
		\end{cases}
		$$ 
		I.e. we remember the input length $n$, and work with the length-$n$ initialization of the $\mathtt{DYCK}-(1,n)$ transformer from our family. 
		\item Similarly on words of length $n$ we use the attention and activation functions of the $\mathtt{DYCK}-(1,n)$-case and leave default values in all other components.
	\end{itemize}
	This transformer works on inputs of length $n$ as a $\mathtt{DYCK}-(1,n)$-recognizing transformer. But a bracket expression of length $n$ can never have depth higher than $n$, i.e. $\mathtt{DYCK}-(1,n) \cap \{(,)\}^n=\mathtt{DYCK}-1 \cap \{(,)\}^n$. But this would mean that our constructed GUHAT transformer would recognize $\mathtt{DYCK}-1$, which is impossible.

\end{proof}

\section{Relations between the transformer variants}

In this section we are going to relate the various classes of transformers (generalized?, masked?, finite image?) that we encountered in the previous section. To have an easier time comparing them, we will rephrase condition distinguishing GUHATs from UHATs in more conceptual terms. For this, we will introduce the notion of separability. We will use this to show that masking in the finite type setting can be simulated by allowing infinite type. Then we will show that in the context of finite images, having masking is more expressive than not having it.

\subsection{Separable attention}
\label{sec:sep}

\begin{definition}\label{def:sep}Let $s:V\times V\to \mathbb R$ be a function. We say that it is separable if there exists a constant $k$ and functions $f_i:V\to \mathbb R$ and  $g_i:V\to \mathbb R$ for $i\in \{1,,,...,k\}$, such that $s(x,y)=\sum_{i=0}^k f_i(x)g_i(y)$.  
\end{definition}

\begin{example}\label{ex:sepalin} The attention function 	
	$$
	s_1((c,i,n),(\tilde c,j, n))=-(n-1-i-j)^2
	$$ from Example \ref{ex:palin} is separable. To express it we set:
	\begin{align*}
		&f_1=-(n-1-i)^2& &g_1=1\\
		&f_2=2(n-1-i) & &g_2=j\\
		&f_3=1  & &g_3=-j^2\\
	\end{align*}
\end{example}

The reason we are interested in separability is that it can replace the (less intrinsic) notion of bilinearity: 

\begin{thm}\label{thm:sep} Languages recognized by a (masked) URASP program with separable attention functions can be recognized by a URASP program with bilinear attention (and the same masking).
\end{thm}

\begin{proof} Let us without loss of generality assume that $V[l]$ are vector spaces for all $l$ (cf. Remark \ref{rem:vspacesimple}).  We will now describe how to change the attention functions to bilinear ones at any given layer $V[l]$ by adding an additional intermediate layer $V'[l]$ before it.
	
Write $V[\leq l]$ for $V[0]\times ...\times V[l]$. Suppose the layer $l$ attention function $s:V[\leq l-1]\times V[\leq l-1]\to \mathbb R$ is separable by  $f_i,g_i:V[\leq l-1]\to \mathbb R$ for $i\in \{1,...,k\}$, such that $s(x,y)=\sum_{i=0}^k f_i(x)g_i(y)$. Then we add an extra layer $V'[l]=\mathbb R^{2k}$ after $V[l-1]$ (but before $V[l]$). Then at this new layer we carry out a pointwise operation which computes the values of $f_i$ and $g_i$ at every position individually. The results are triples $v=(v_o,v_f,v_g)$ in $V[\leq l-1]\times \mathbb R^{k}\times \mathbb R^{k}$ Then from $V'[l]$ to the $V[l]$ we pick as attention function the bilinear function 
$$a( (v_o,v_f,v_g),(w_o,w_f,w_g) )=\langle v_f,v_g\rangle.$$
\end{proof}

\begin{remark}~
	\begin{itemize}
		\item We note that the constant $k$ is supposed to be independent of $n$. If it were to depend on $n$, then the necessary size of the cells $V'[d]$ in the new transformer might have to increase with $n$, which is not allowed.
		\item In the proof above we are doubling the lines of the program (or layers), but this is just to simplify exposition. Any program with $L_p+L_a$ layers, where $L_p$ are point-wise layers and $L_a$ are actual attention layers, can be realized in $L_a+1$ layers, by absorbing all pointwise computations into the activation of the attention layers preceeding them.
	\end{itemize}
	
\end{remark}

\begin{cor}\label{cor:pal} The language $\mathtt{PALINDROMES}$ can be realized by a UHAT or a MUHAT. This has been observed in \cite{kozachinskiy2023logical}, and is a consequence of the above theorem and Examples \ref{ex:mpalin}, \ref{ex:palin} and \ref{ex:sepalin}.
\end{cor}

\begin{lem}\label{lem:finitelyalliseparable} Let $V$ be a finite set and $s:V\times V\to \mathbb R$ any function. Then $s$ is separable.
\end{lem}
\begin{proof} Let $V=\{v_1,...,v_l\}$ and set $s_{i,j}=s(v_i,v_j)$. Then we take $k=l^2$ and set for $\alpha,\beta\in \{1,...,l\}$
	\begin{align*}
		f_{\alpha l+\beta}(v_i)=\begin{cases}
			s_{i,\beta} & \mathrm{if~} i=\alpha\\
			0 & \mathrm{else}
		\end{cases}, && 
		g_{\alpha l+\beta}(v_j)=\begin{cases}
			1 & \mathrm{if~} j=\beta\\
			0 & \mathrm{else}
		\end{cases},
	\end{align*}
	This way the for any pairing of these only one coordinate in the scalar product is non-zero and has exactly the value $s_{i,j}$.
\end{proof}

A direct consequence is that the difference between GUHAT and UHAT disappears in the finite case:

\begin{cor} \label{cor:finautosep} Any language recognizable by a finite (masked) GUHAT can be recognized by a finite UHAT (with the same masking).
\end{cor}

\subsection{Simulating masking}

No inclusion relation between \emph{UHAT} and \emph{MUHAT} is known. The trick which we used to simulate masking in the GUHAT setting does not preserve separability. However, when the attention values are binary, there is a way out. This binarity can always be achieved in the setting of finite type transformers.

\begin{lem}\label{lem:binarize} Any language in \emph{F-MUHAT}, can also be described by a finite image transformer with strict future-masking masking where the attention functions can only take the values $\{0,1\}$.
\end{lem}

\begin{proof} 
This is a consequence of the results of \cite{yangMaskedHardAttentionTransformers2024}. There an equivalence between \emph{F-MUHAT} and languages recognizable by a programming language called \emph{BRASP} (boolean RASP) is established. \emph{BRASP} is like URASP with masking, but with two differences:
\begin{itemize}
	\item all $V[i]$ are binary, i.e. equal to $\{0,1\}$ and all functions involved (attention functions $s$, activation functions $v$ are binary valued).
	\item the initialization may use more than the first layer for encoding the characters of the string and their positional encoding. (since you can not encode much into one bit). 
\end{itemize}
In particular, given a F-MUHAT language, we can pick a BRASP program recognizing it and interpret it as a URASP program by collapsing all initialization layers into one non-binary layer and leaving everything else as is. This will give us a URASP program with the desired properties.
\end{proof}
The above observation allows us to simulate $F-MUHAT$ by $UHAT$. 

\begin{thm}[\cite{barceloLogicalLanguagesAccepted2023,yangMaskedHardAttentionTransformers2024}]\label{thm:finitemaskingtouhat} \emph{F-MUHAT} $\subset$ \emph{UHAT}.
\end{thm}

The proof of the theorem follows by combining the results of \cite{barceloLogicalLanguagesAccepted2023} and \cite{yangMaskedHardAttentionTransformers2024} (cf. Theorems \ref{thm:fomoninuhat} and \ref{thm:fomonisfmguhat}). However, using the notion of separability we can obtain a slightly more direct proof:

\begin{proof}We start with an $F-MUHAT$-transformer, respectively its corresponding URASP program. By Lemma \ref{lem:binarize}, we can assume it to have attention functions $s_l$ valued in $\{0,1\}$. Since we are in the finite setting, these are separable (Lemma \ref{lem:finitelyalliseparable}). We claim that by adding the current position $i$ and the total length $n$ to the positional encoding, there exist attention functions $\bar s_l$ with the following properties:
	\begin{enumerate}
		\item $\bar s_l$ is separable
		\item $\bar s_l((v,i,n),(w,j,n))< 0$ when $j\geq i$
		\item $\bar s_l((v,i,n),(w,j,n))\in (0,\frac{1}{3})$ when $j<i$ and $s_l(v,w)=0$
		\item $\bar s_l((v,i,n),(w,j,n))>\frac{1}{2}$ when $j<i$ and $s_l(v,w)=1$
		\item  $\bar s_l((v,i,n),(w,j,n))<\bar s_l((v,i,n),(w',j',n))$ when $j<j'<i$ and $s(v,w)=s(v,w')$.
	\end{enumerate}
	If we have such an attention function, then the proof is finished, by adding $(i,n)$ to the positional encoding and replacing the attention functions $s_l$ by $\bar s_l$. The values after (or equaling) $i$ will always be negative and the ones before $i$ always positive, so the 'masking' works (except for the very first position, where the default value should be used). The values which used to have attention zero are all smaller than the others, hence they will only be attended to if no 1-value exists. Moreover the 5. property guarantees that 'rightmost-hardness' is preserved.

	So let us define $\bar s_l$ and show its properties:
	$$
	s_l((v,i,n),(w,j,n)):= \left(i-j-\frac{1}{2}\right)e^{3j}\cdot \left(s_l(v,w)+ \frac{1}{4 ne^{3n}}\right)
	$$
	First of all let us note that the first term $\left(i-j-\frac{1}{2}\right)$ determines the sign of the whole expression, and is negative exactly when $j>i$. Hence it satisfies property 2. For the monotonicity property 5 it suffices to verify that the derivative of the function with respect to the $j$-variable is strictly positive for constant $s_l(v,w)$. We calculate only with the non-constant term:
	\begin{align*}
		&\frac{\partial}{\partial j}\left(\left(i-j-\frac{1}{2}\right)e^{3j}\right)
		&=3\left(i-j-\frac{1}{2}\right)e^{3j} - e^{3j}
		&=3\left(i-j-\frac{1}{2}-\frac{1}{3}\right)e^{3j}
		&=3\left(i-j-\frac{5}{6}\right)e^{3j} 
	\end{align*}
	For $j<i$ this is strictly positive, hence property 5 is satisfied. In order to verify the properties 3 and 4, let us look at what values the function can take when $j<i$: The first term $(i-j-\frac{1}{2})$ is always at least $\frac{1}{2}$ and certainly smaller than $n$. The second term $e^{3j}$ is at least 1 and always smaller than $e^{3n}$. In particular, when $s_l(v,w)=1$, the total expression is certainly $>\frac{1}{2}$ as asserted by property 4. On the other hand, the product of the first terms is always smaller than $ne^{3n}$, hence when $s_{l}(v,w)=0$, the total value of the modified function is at most:
	$$
	\frac{ne^{3n}}{4 ne^{3n}}<\frac{1}{3}
	$$
	as desired by property 3. What remains to show is Property 1, i.e. that the attention function $\bar s_l$ is separable. This is guaranteed, because finite sums and products of separable functions are separable, as asserted by the Lemma below.
\end{proof}

\begin{lem} \label{lem:sepisalg}Finite sums and products of separable attention functions are separable.
\end{lem}

\begin{proof} The short algebraic reason is that formally speaking a separable attention function is a function $V[l]\times V[l]\to \mathbb R$ induced by an element of $Map(V[l],\mathbb R)\otimes Map(V[l],\mathbb R)$, where $Map(V[l],\mathbb R)$ denotes the space of all functions from $V[l]$ to $\mathbb R$, and $\otimes$ denotes the usual tensor product of rings. Now this tensor product of rings is a sub-ring of $Map(V[l]\times V[l],\mathbb R)$, i.e. stable under finite sums and products. Let us make this a little more explicit, using the notations from Definition \ref{def:sep}:
	Given two elements: $\sum_{i=1}^k f_k\otimes g_k$ and  $\sum_{i=1}^{k'} f_k'\otimes g_k'$, which represent the separable attention functions $a(v,w)=\sum_{i=1}^kf_i(v)g_i(w)$ respectively  $a'(v,w)=\sum_{i=1}^{k'}f_i'(v)g_i'(w)$, their sum $\bar a=a+a'$ is given by 
	$
	\sum_{i=1}^{k+k'}\bar f_i\otimes \bar g_i, 
	$
	where
	\begin{align*}
		\bar f_i=\begin{cases}
			f_i & \mathrm{when~}i\leq k\\
			f_{i-k}' & \mathrm{else}
		\end{cases} &&
		\bar g_i=\begin{cases}
			g_i & \mathrm{when~}i\leq k\\
			g_{i-k}' & \mathrm{else}.
		\end{cases} 
	\end{align*}
	Similarly $\tilde a=a\cdot a'$ can be expressed as
	$
	\sum_{i=1}^{k\cdot k'}\tilde f_i\otimes \tilde g_i, 
	$
	where we set for $i=\alpha + \beta k$ with $\alpha\in \{1,...,k\}$ and $\beta\in \{1,...,k'\}$
	$$
	\tilde f_i=f_\alpha\cdot f_\beta' \mathrm{~~~and~~~} \tilde g_i=g_\alpha\cdot g_\beta' .
	$$
\end{proof}

\subsection{Masking improves expressibility when the type is finite}

For GUHAT-type transformers, where the image of the positional encoding is not limited, masking was (potentially) a defect: By Remark \ref{rk:maskedguhat} an unmasked GUHAT can do everything a masked one does. When we require finiteness, this relation is reversed: A masked finite image transformer can recognize everything an unmasked one can, and even more. Let us first prove the inclusion and then provide an explanation of strictness of the inclusion:

\begin{thm} \label{thm:fmaskedbetter} F-UHAT $\subset$ F-MUHAT.
\end{thm}

\begin{proof} Let a language in \emph{F-UHAT} be given. 
	\begin{itemize}
		\item By \cite[Theorem 4]{yangMaskedHardAttentionTransformers2024}, we can translate  it to a BRASP program without masking.
		\item By following the procedure from the proof of \cite[Proposition 11 and Lemma 2]{yangMaskedHardAttentionTransformers2024}, we can change the BRASP program to one, where the attention and activation of all attention operations only depends on the index attended to, not the current position, i.e. are of the form
		\begin{align}\label{eq:brasplineunmaseked}
			\blacktriangleleft_j[\top,s(j)]v(j):D(i)
		\end{align}
		where $\top=1$ corresponds to the 'always true' masking, i.e. no masking. 
		We note that since there is no masking, $D(i)$ is never used, i.e. we might set it to zero and forget about it.
		\item Now we can replace any unmasked attention layer like \eqref{eq:brasplineunmaseked} by the following sequence of four layers, where the attention mechanisms use masking.\footnote{Note that here we use both future-masking with rightmost-hard attention and past-masking with leftmost hard attention.}
		\begin{align*}
			&BOS(i)&=&\blacktriangleright_j[i>j, 0] 0: 1\\
			&L_1(i)&=&\blacktriangleleft_j[i<j, s(j)] v(j): 0\\
			&L_2(i)&=&s(i)\wedge v(i) \vee \neg s(i)\wedge L_1(i)\\ 
			&L_3(i)&=&\blacktriangleright_j[i>j,BOS(j)] L_2(j): L_2(i) 
		\end{align*} 
		Here BOS (beginning of sequence) is $1$ only for the first position (where the strict masking forces us to use the default value) and $0$ at all other positions. In the position $i=0$, the second line $L_1$ computes the leftmost $v(j)$ with $s(j)=1$, we don't really care what it does in the positions. $L_2$ is a point-wise operation, repairing the fact that since we had strict masking $L_1(0)$ does not take into consideration whether $s(0)=1$. For $L_2$ we are also only interested in the first position. Finally $L_3$ is transporting the result of $L_2$ to all positions.
		
		This way we can replace our unmasked BRASP program with one using only (strictly future and past) masked attention operations.
		\item  The result now follows from Remark \ref{rem:pastfutre}.
	\end{itemize}
\end{proof}

We will now see that the inclusion of languages recognizable by unmasked finite image GUHATS into strictly masked ones is strict. For this we will use the following example (which is an adapted version of an argument after Corollary 9 in \cite{yangMaskedHardAttentionTransformers2024}): 

\begin{example} \label{ex:needmask} Recall that the language $\mathtt{DYCK}-(1,2)$ is star-free, hence in $FO_<$, i.e. recognizable by a $F-M(G)UHAT$. However, it is not recognizable by $F-(G)UHAT$ transformers by the following reasoning:
	
	Assume there was an  $F-(G)UHAT$ recognizing it. Since the positional encoding (resp. space of predicates $Q$) is finite, there is some big $n$ such that there exist numbers $i<j<k$ with identical positional encoding (on $n$-character strings). 
	
	One can construct a valid  $\mathtt{DYCK}-(1,2)$ expression which has an opening brace at $i$ and a closing brace at $j$. Whatever the value of the string of position $k$, we know that changing it makes the string an invalid expression, but we claim that no $F-(G)UHAT$ can recognize this change:
	
	To see this, we translate the F-(G)UHAT into a BRASP program and use the same technique as in Theorem \ref{thm:fmaskedbetter} to have all attention operations have the shape as in Equation \eqref{eq:brasplineunmaseked}. Since the column at $k$ will always coincide with the column at $i$ or $j$ (which are preferred to it in the leftmost-hard case), the $BRASP$ program will operate identically on the two inputs, even though one is in the language and the other isn't.
\end{example}

\subsection{Summary}

The classes of languages we have discussed thus far fit into the following diagram of inclusions:
\[\begin{tikzcd}
	&&&& {AC^0} \\
	&&&& \begin{array}{c} AC^0\cap \substack{\mathrm{strongly} \\\epsilon-\mathrm{fixable}} \end{array} \\
	& MGUHAT &&& GUHAT \\
	\\
	& MUHAT &&& UHAT \\
	\\
	{F-MGUHAT} & {F-MUHAT} &&& {F-UHAT} & {F-GUHAT} \\
	& {FO_{<}(Mon)}
	\arrow["{\neq }"', hook, from=2-5, to=1-5]
	\arrow[hook, from=3-2, to=3-5]
	\arrow[hook, from=3-5, to=2-5]
	\arrow[hook, from=5-2, to=3-2]
	\arrow[hook, from=5-5, to=3-5]
	\arrow[equals, from=7-1, to=7-2]
	\arrow["\neq", hook, from=7-2, to=5-2]
	\arrow["\neq", hook, from=7-2, to=5-5]
	\arrow[equals, from=7-2, to=8-2]
	\arrow["\neq", hook, from=7-5, to=5-5]
	\arrow["\neq"', hook', from=7-5, to=7-2]
	\arrow[equals, from=7-5, to=7-6]
\end{tikzcd}\]

In the diagram the following classes of languages are used:
\begin{itemize}

\item The definition of strong $\epsilon$-fixability is provided in Appendix \ref{def:strongeps}.
\item $AC^0$ stands for the space of languages that can be recognized by constant depth and polynomial size circuit families (cf. Definition \ref{def:Ac0}).
\item $FO_<(Mon)$ is the class of first order languages using only unary numerical predicates and the binary numerical predicate $<$, cf. Definitions \ref{def:foq} and \ref{def:mon}.

\end{itemize}

Let us explain the individual inclusions:

\begin{itemize}
\item The following inclusions are consequences of the definition (getting less restrictive from left to right): 
	\begin{align*}
		&	F-UHAT&&\to &&UHAT&&\to &&GUHAT\\
		\mathrm{and~~~}&F-MUHAT&&\to &&MUHAT&&\to &&MGUHAT
	\end{align*}
\item The characterization of $F-M(G)UHAT$ by first order logic with unary numerical predicates $FO(Mon)$ has been done in \cite{yangMaskedHardAttentionTransformers2024}, cf. Theorem \ref{thm:fomonisfmguhat}.
\item The inclusion $GUHAT\to AC^0\cap\substack{\mathrm{strongly} \\\epsilon-\mathrm{fixable}}$ is a combination of the results of \cite{hahn-2020-theoretical} and \cite{haoFormalLanguageRecognition2022}, cf. Theorems  \ref{thm:strongfix} and \ref{thm:guhatinac0}.
\item The inclusion $
AC^0\cap\substack{\mathrm{strongly} \\\epsilon-\mathrm{fixable}}
\to AC^0$ was first observed in \cite{barceloLogicalLanguagesAccepted2023} and is based on the language $\mathtt{APPROXIMATE}$-$\mathtt{MAJORITY}$  which lies in $AC^0$, but by construction can not be (strongly) $\epsilon$-fixable (cf. Theorem \ref{thm:approxmaj} and Corollary \ref{cor:approxmaj}).
	
	\item The inclusion $MGUHAT \to GUHAT$ follows from Remark \ref{rk:maskedguhat}.

	\item The equalities $F-GUHAT=F-UHAT$ and $F-MGUHAT=F-MUHAT$ were proven in Corollary \ref{cor:finautosep}).
		
	\item The diagonal inclusion $F-M(G)UHAT\to UHAT$ follows from the results of \cite{barceloLogicalLanguagesAccepted2023} and \cite{yangMaskedHardAttentionTransformers2024} combined (cf. Theorems \ref{thm:fomonisfmguhat} and \ref{thm:fomoninuhat}). It can also be proven more directly using \ref{thm:finitemaskingtouhat}.
	
	\item The inclusion $F-(G)UHAT\to F-M(G)UHAT$ was established in Theorem \ref{thm:fmaskedbetter} and its strictness follows from Example \ref{ex:needmask}.

	\item The inclusions $F-(G)UHAT\to UHAT$, $F-M(G)UHAT\to MUHAT$ and $F-M(G)UHAT\to UHAT$ are strict, since the finite image languages can not recognize $\mathtt{PALINDROMES}$ (Theorem \ref{thm:nopal}) while both UHAT and MUHAT do (Corollary \ref{cor:pal}).

\end{itemize}

The following inclusions in the above diagram seem open:

\begin{open}~
	\begin{itemize}
	\item Can any language recognizable by GUHAT, also be recognized by a masked GUHAT?
\item Is there a language recognizable by (masked) GUHAT, which can not be recognized by any (masked) UHAT?
		\item Can any language recognizable by a UHAT be recognized by a masked UHAT?
		\item Can any language recognizable by a  future-masked UHAT be recognized by a UHAT?
	\end{itemize}
\end{open}

\appendix

\section{Background on relevant language classes and results}
\subsection{strong $\epsilon$-fixability}
In \cite{hahn-2020-theoretical}, it is shown that languages recognized by GUHAT transformers are strongly $\epsilon$-fixable.
Strong $\epsilon$-fixability is the fact that by imposing an arbitrarily small fraction of the characters, we can determine whether the string does or does not belong to the language. Formally it is defined as follows:

\begin{definition} \label{def:strongeps} Let $\Sigma$ be an alphabet and $?$ a symbol not in $\Sigma$. We call \emph{restriction} of $\Sigma^*$ any element $\rho$ of $(\Sigma\cup \{?\})^*$. We think of $?$ to be a wildcard which can be replaced by any character of $\Sigma$. An evaluation of $\rho$ is an element of $\Sigma^*$ which coincides with $\rho$ on the non-$?$-characters.\\   
We will write $|\rho|$ for the number of non-wildcard characters in a restriction. We will say that a restriction $\rho'$ extends $\rho$, writing $\rho \succcurlyeq \rho'$, if $\rho'$ is obtained from $\rho$ by fixing wildcards to characters of $\Sigma$. \\
We say that a language $L$ is \emph{strongly $\epsilon$-fixable} if for all $\epsilon>0$, there is an $N\in\mathbb N$, such that for all $n>N$ and restrictions $\rho$, there is a an extension $\rho'$ of $\rho$ with $|\rho'|\leq |\rho|+\epsilon n$, such that  either
\begin{itemize}		
	\item all evaluations of $\rho$ are in $L$,
	\item or no evaluation of $\rho$ is in $L$.
\end{itemize}
\end{definition}

\begin{thm}[\cite{hahn-2020-theoretical}]\label{thm:strongfix} Any language in \emph{GUHAT} is strongly $\epsilon$-fixable.
\end{thm}

\begin{remark}The theorem is stated there for $\epsilon$-fixability (which would always start with the arbitrary restriction $?^n$), but actually proven (and used) in the stronger case. Moreover $\Sigma$ is assumed to be $\{0,1\}$ in the paper (a condition that can be dropped), however multiple attention heads are allowed (which does not increase expressivity). 
\end{remark}

The relevance of this theorem is that it allows us to see that GUHAT transformers can not recognize all languages in $AC^0$. We follow \cite{barceloLogicalLanguagesAccepted2023} in this observation and need the following theorem:

\begin{thm}[\cite{ajtai1984probabilistic} cf. also \cite{goldreich_approx_majority} for a short self-contained proof] \label{thm:approxmaj} Let $\Sigma=\{0,1\}$. For $w=w_1...w_n\in \Sigma^n$, we write $|w|=\sum_{i=1}^n w_i$ and call this value the Hamming weight of $w$.\\
	There is a language over $\Sigma=\{0,1\}$ in $AC^0$ with the following properties for every $n\in \mathbb N$:
	\begin{itemize}
		\item If $w\in \Sigma^n$ satisfies $|w|\leq \frac{n}{3}$, then $w\not\in L$.
		\item If $w\in \Sigma^n$ satisfies $|w|\geq \frac{2n}{3}$, then $w \in L$. 
	\end{itemize}
\end{thm}

We will call any language in $AC^0$ with this property $\mathtt{APPROXIMATE~MAJORITY}$. There are many some languages, and also there are many languages with the above property not lying in $AC^0$. 

\begin{cor}[\cite{barceloLogicalLanguagesAccepted2023}]\label{cor:approxmaj} $\mathtt{APPROXIMATE~MAJORITY}$ is not strongly $\epsilon$-fixable.
\end{cor}

\subsection{Circuit complexity and the class $AC^0$}
\label{sec:1circ}
In this section we recall the class $AC^0$ of formal languages accepted by a family of Boolean circuits of constant depth and polynomial size.

\begin{definition}A boolean circuit with $n$ inputs and $m$ outputs is given by a tuple $$(V,E,L,z_1,...,z_m)$$ where
	\begin{itemize}
		\item $(V,E)$ is a directed acyclic graph, we call its edges \emph{wires}.
		\item $L:V\to \{x_1,...x_n, \mathbf{0}, \mathbf{1},NOT, AND, OR\}$ is a labeling function. such that:
		\begin{itemize}
			\item The $x_i$ have fan-in 0, each of them occurs exactly once. We call them \emph{input vertices}, all other vertices are called \emph{gates}.
			\item $\mathbf{0},\mathbf{1}$ have fan-in 0.
			\item $NOT$ has fan-in 1.
			\item $AND, OR$ have arbitrary fan-in.
		\end{itemize}
		\item $z_1,...,z_m\in V$ are not necessarily distinct output vertices.
	\end{itemize}
\end{definition}

\begin{definition}A family of circuits $C=\{C_n, n\in \mathbb N\}$ is a circuit $C_n$ with $n$ inputs and one output for any number $n$. We say that:
	\begin{itemize}
		\item $C$ has constant depth, if there exists a constant $K$ such that the depth of $C_n$ is bounded by $K$ for all $n$. The depth of a graph denotes the length of the longest path in it.
		\item  $C$ has polynomial size, if there exists a constant $c$ such that the number of wires of $C_n$ is less than $n^c+c$ for all $n$.
	\end{itemize}
	We denote by $AC^0$ the class of families of circuits of constant depth and polynomial size.
\end{definition}

By using the usual circuit arithmetic and applying $C_n$ to elements of $\{0,1\}^n$, such a family of circuits determines a function $f_C:\{0,1\}^*\to \{0,1\}$, i.e. a boolean function on $\{0,1\}^*$.

We would like to be able to talk about the $AC^0$ class also for languages over other alphabets and we can do so, by composing $f_C$ with a binary symbol encoding, i.e. an injective map $h:\Sigma\to \{0,1\}^s$ (where $s=\lceil\log_2(|\Sigma|)\rceil$). By element-wise application and concatenation such a map $h$ induces a map $\bar h:\Sigma^*\to \{0,1\}^*$.
\begin{definition}\label{def:Ac0}
	We say that a language $L$ over $\Sigma$ is in  $AC^0$ if there is a binary symbol encoding $h$ and $AC^0$ family of circuits $C$ such that for $w\in \Sigma^*$ we have $w\in L$ if and only if $f_C\circ \bar h(w)=1$.
\end{definition}

We can now formulate:
\begin{thm}[Theorem 1 in \cite{haoFormalLanguageRecognition2022}]\label{thm:guhatinac0} Any language in \emph{GUHAT} is in $AC^0$.
\end{thm}

We also quote the following observation:

\begin{cor}[\cite{haoFormalLanguageRecognition2022}] \label{cor:dycknotac}The language $\mathtt{DYCK}-1$ for correct bracket expressions over the alphabet $\Sigma=\{(,)\}$ is not in $AC^0$ and hence not in GUHAT.
\end{cor}

\subsection{First-order formulas}

We now want to introduce the class of languages which can be defined by first order logic formulas.

\begin{definition} \label{def:foq}
	We call $FO_{<}$ the class of languages which can be defined by sentences in first order logic (using the binary numerical predicate  $<$). Let $Q$ be a class of predicates, then we call $FO_{<}(Q)$ the class of languages which one obtains by adding $Q$ to the space of elementary predicates.  
\end{definition}

We will be primarily interested in the following class of predicates: 

\begin{definition}\label{def:mon} A unary numerical (or monadic) predicate $\theta$ is a collection of functions $\theta_n:\{0,...,n-1\}\to \{0,1\}$ for $n\in\mathbb N$. We write $Mon$ for this class of predicates. 
\end{definition}

An equivalent way to express the same class of languages is linear temporal logic. Let us start with  future linear temporal logic ($FLTL$). In addition to elementary predicates $P:\Sigma\to \{\top,\bot\}$ (and potentially an additional set of predicates $Q$) and Boolean connectors  $\neg$, $\vee$, $\wedge$ between sentences, these are built from:
\begin{itemize}
	\item For any sentence $\phi$, the sentence $X\phi$ - read 'neXt $\phi$'
	\item For any sentences $\phi, \psi$ the sentence $\phi U\psi$ - read '$\phi$ Until $\psi$'
\end{itemize}

The way we check for $w=(a_0,...,a_{n-1})$ whether it accepted by $\phi$ uses the satisfiability $(w,i)\vDash \phi$ of $w$ at position $i$. The satisfaction is given by the following rules:
\begin{itemize}
	\item $(w,i)\vDash P$ if $P(w_i)$.
	\item boolean operators work as usual.
	\item $(w,i)\vDash X\phi$ if $i<n-1$ and $(w,i+1)\vDash \phi$
	\item  $(w,i)\vDash \phi U \psi$ if $\exists j\in\{i+1,...,n-1\}$ such that $(w,j)\vDash \psi$ and $(w,k)\vDash \phi$ for $j\in\{i+1,...,j-1\}$.
\end{itemize}

We say that a word $w$ is in the language defined by $\phi$ in $FLTL$ if $(w,0)\vDash \phi$.

\begin{definition} 
	We call $FLTL$ (future linear temporal logic) the class of languages which can be defined by sentences in $FLTL$. Let $Q$ be a class of predicates, then we call $FLTL(Q)$ the class of languages which one obtains by adding $Q$ to the space of elementary predicates.  
\end{definition}

It is clear that $X\phi$ can be expressed by $\bot U \phi$. What is less clear is that adding 'past analogues' to the future operators does not increase expressivity. A sentence as above, but instead of $X$ and $U$ using the operators:
\begin{itemize}
	\item $Y$ (Yesterday). $(w,i)\vDash X\phi$ if $0<i$ and $(w,i-1)\vDash \phi$
	\item $S$ (Since).   $(w,i)\vDash \phi S \psi$ if $\exists j\in\{0,...,i-1\}$ such that $(w,j)\vDash \psi$ and $(w,k)\vDash \phi$ for $j\in\{j+1,...,i-1\}$.
\end{itemize} 
will be called a $PLTL$ (Past linear temporal logic) sentence. For a $PLTL$ sentence $\phi$, we will say that $w\in \Sigma^n$ is in the language if $(w,n-1)\vDash \phi$. We call $LTL$ the languages allowing both the past and future operators. We will need the following theorem that asserts that all the descriptions of this section (and in fact many more) are equivalent:

\begin{thm}[ \cite{diekertFirstorderDefinableLanguages2008}]~\label{thm:foltlstarfree}
	$FO_<(Q)=LTL(Q)=PLTL(Q)=FLTL(Q)$
\end{thm}

We note that the compatibility with the additional predicates is not very explicit in \cite{diekertFirstorderDefinableLanguages2008}, however it is noted in \cite[Propsition 2]{barceloLogicalLanguagesAccepted2023}. Similarly, the equivalence of $LTL$, $PLTL$ and $FLTL$ is stated more explicitly e.g. in \cite{Markey2003TemporalLW}.\\

This equivalence was shown to prove that:

\begin{thm}[\cite{barceloLogicalLanguagesAccepted2023}]\label{thm:fomoninuhat} The languages in $FO_<(Mon)$ are contained in \emph{UHAT}.
\end{thm}

The inclusion in this theorem is strict, as explained in the talk \cite{kozachinskiy2023logical}, where a MUHAT recognizing $\mathtt{PALINDROMES}$ is presented along with the following observation:

\begin{thm} \label{thm:nopal} The language $\mathtt{PALINDROMES}$ is not in $FO(Mon)$.
\end{thm}
The proof of this using Ehrenfeucht-Fraissé games was kindly explained to us by Corentin Barloy.\\

In the context of masked transformers with finite image, the following equivalence was established:

\begin{thm}[\cite{yangMaskedHardAttentionTransformers2024}]\label{thm:fomonisfmguhat} The languages in $F-MUHAT$ are exactly $FO_<(Mon)=LTL(Mon)$. 
\end{thm}

\begin{remark} \label{rem:pastfutre}The proof of this theorem translates PLTL formulas into BRASP programs (cf. proof of Lemma \ref{lem:binarize} for Remarks on BRASP). However, the translation mechanism can also translate between LTL formulas (involving the next/until operations) to slightly more general BRASP programs, which can also use $\blacktriangleleft$ (past-masked with leftmost hard tie resolution) attention layers. Due to Theorem \ref{thm:foltlstarfree}, this means that BRASP programs involving both $\blacktriangleleft$ and $\blacktriangleright$ layers can express exactly the same languages as those only using $\blacktriangleright$-layers.
\end{remark}

\begin{example}\label{ex:dyck}
The language $\mathtt{DYCK}-(1,D)$, of bracket expressions of depth $\leq D$ can be computed by an $F-MUHAT$ (or $UHAT$) transformer for any $D$. This is a consequence of  $\mathtt{DYCK}-(1,D)$ being star-free \cite{BRZOZOWSKI197837}, and hence in LTL (cf. e.g. \cite{diekertFirstorderDefinableLanguages2008}).
\end{example}

\bibliographystyle{plain}
\bibliography{bibliography}


\end{document}